\documentclass[10pt,twocolumn,letterpaper]{article}

\usepackage[final]{cvpr}              

\usepackage[dvipsnames]{xcolor}

\usepackage{graphicx}
\usepackage{multirow}
\usepackage{algorithm}
\usepackage{algorithmic}

\definecolor{cvprblue}{rgb}{0.21,0.49,0.74}
\usepackage[pagebackref,breaklinks,colorlinks,citecolor=cvprblue]{hyperref}

\def\paperID{*****} 
\def\confName{CVPR}
\def\confYear{2024}

\title{FedProK: Trustworthy Federated Class-Incremental Learning \\via Prototypical Feature Knowledge Transfer}

\author{Xin Gao$^1$\footnote[1], \quad Xin Yang$^1$\footnote[2],\quad Hao Yu$^1$\footnote[1],\quad Yan Kang$^2$,\quad Tianrui Li$^3$
\\
$^1$Southwestern University of Finance and Economics,\quad $^2$Webank,\quad $^3$Southwest Jiaotong University
\\ 
{\tt{\small{xingaocs@foxmail.com, yangxin@swufe.edu.cn,}}}
\\ 
{\tt\small yuhao2033@163.com, yangkang@webank.com, trli@swjtu.edu.cn}}

\begin{document}
\maketitle
\renewcommand{\thefootnote}{*}
\footnotetext[1]{Equal contribution} 
\renewcommand{\thefootnote}{†}
\footnotetext[2]{Corresponding author}

\begin{abstract}
{Federated Class-Incremental Learning (FCIL) focuses on continually transferring the previous knowledge to learn new classes in dynamic Federated Learning (FL). However, existing methods do not consider the trustworthiness of FCIL, i.e., improving continual utility, privacy, and efficiency simultaneously, which is greatly influenced by catastrophic forgetting and data heterogeneity among clients. To address this issue, we propose FedProK (\textbf{Fed}erated \textbf{Pro}totypical Feature \textbf{K}nowledge Transfer), leveraging prototypical feature as a novel representation of knowledge to perform spatial-temporal knowledge transfer. Specifically, FedProK consists of two components: (1) feature translation procedure on the client side by temporal knowledge transfer from the learned classes and (2) prototypical knowledge fusion on the server side by spatial knowledge transfer among clients. Extensive experiments conducted in both synchronous and asynchronous settings demonstrate that our FedProK outperforms the other state-of-the-art methods in three perspectives of trustworthiness, validating its effectiveness in selectively transferring spatial-temporal knowledge.}
\end{abstract}   
\section{Introduction}
\label{sec:intro}

Artificial Intelligence (AI) has profoundly impacted humanity in various aspects \cite{teeti2022vision,feng2022legal}. However, with the increasing awareness of ethics, fairness, and security, the centralized training and data storage paradigm in traditional deep learning has been greatly challenged \cite{mothukuri2021survey,abdulrahman2020survey}. The academic and industrial communities call for the establishment of trustworthy AI systems.

\begin{figure}[t]
  \centering
  \includegraphics[width=0.48\textwidth]{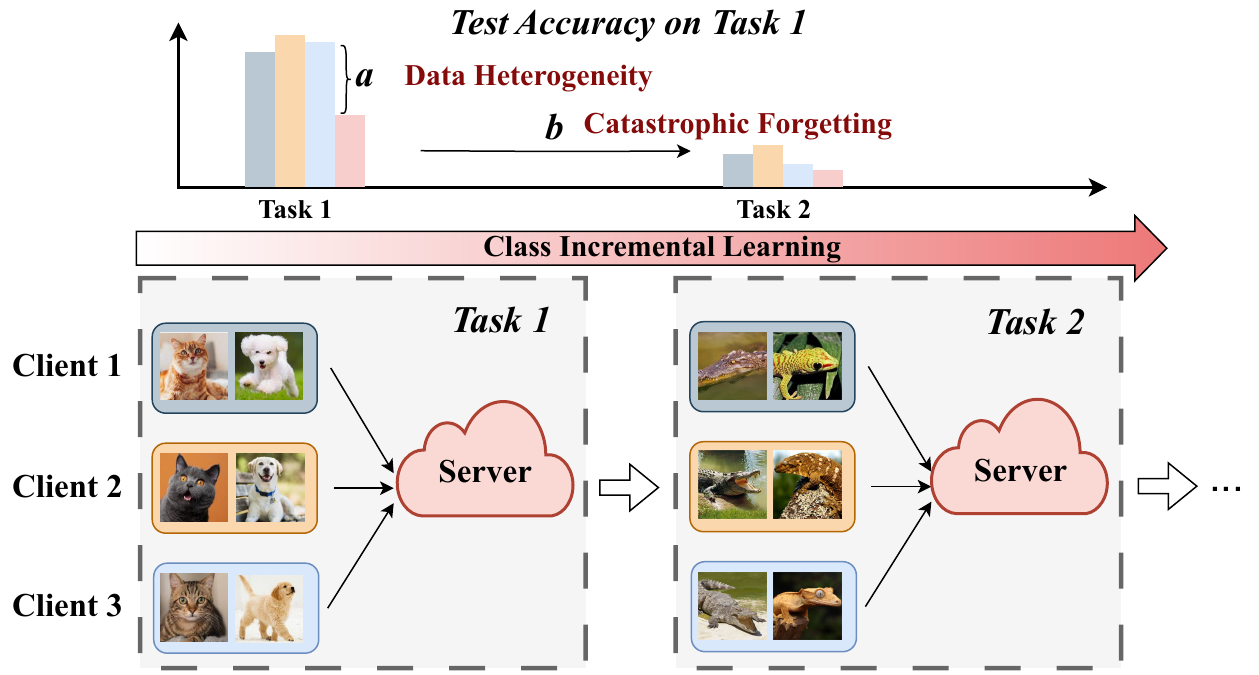}
  \caption{Illustration of the catastrophic forgetting and data heterogeneity in FCIL. The bar chart shows the test accuracy when vanilla FedAvg is employed in this scenario. $a$: The heterogeneous local data leads to poor performance of the global model. $b$: After the completion of training on Task 2, the model suffers catastrophic forgetting of Task 1.}
  \label{fig_intro}
\end{figure}
Federated Class Incremental Learning (FCIL) is a dynamic and heterogeneous paradigm of Federated Learning (FL) \cite{mcmahan2017communication,liu2022vertical}, depicting a more realistic scenario of iterative federated training: data reaches clients sequentially in the form of task streams. While its privacy preservation and adaptivity to non-stationary environments make FCIL a promising solution for trustworthy AI, it is faced with two key challenges. The first challenge is the \textbf{catastrophic forgetting} of the previous knowledge \cite{yang2023federated}. After completing the training on new tasks, models tend to forget the knowledge of previous tasks due to the overwriting of parameters \cite{dong2022federated,yoon2021federated,zhang2023target}. Another key challenge is \textbf{data heterogeneity} among clients, which causes severe performance degradation and unstable convergence of the global model \cite{zhang2021survey,mothukuri2021survey,lyu2020threats}. By introducing diverse and varying data distributions across clients, data heterogeneity significantly exacerbates catastrophic forgetting \cite{chai2023survey}. As illustrated in \cref{fig_intro}, the intertwined challenges of FCIL require more effective approaches to knowledge transfer.

However, existing FCIL methods exhibit certain limitations in addressing these issues. A typical approach of FCIL is to store or generate exemplars of previous classes and replay them when training on new data \cite{dong2022federated,dai2023fedgp,zhang2023target}. However, the performance is heavily influenced by the size of the exemplar set, and the continually growing number of exemplars may exceed the memory capacity of edge devices \cite{dong2022federated}. \cite{zhang2023target} generates pseudo samples on the server and distributes them to the clients, which greatly increases the communication cost of federated training and vulnerability of potential attacks. Former works on traditional centralized continual learning (CL) emphasized that a capable CL model should avoid overfitting to previous tasks when combating catastrophic forgetting \cite{Wu_2021_ICCV,kim2023stability,zhao2020maintaining}. Instead, a capable CL model should strike a delicate balance: preserving previous knowledge (stability) while accommodating new knowledge (plasticity), which is completely overlooked in current FCIL research \cite{yang2023federated}. Besides, simply integrating continual learning methods into the federated setting demonstrates poor performance \cite{ma2022continual}, which is tested in our experiments. 

Therefore, we formulate a trustworthy FCIL framework to evaluate the effectiveness of spatial-temporal knowledge transfer, which comprises three main aspects: \textbf{continual utility}, \textbf{efficiency}, and \textbf{privacy}. Continual utility represents the performance of the global model on overcoming forgetting and data heterogeneity, which can be tested with the final accuracy of the global model and the trade-off of the stability-plasticity dilemma. To enhance privacy preservation, FCIL algorithms should avoid data leakage in addressing catastrophic forgetting. As for efficiency, communicational cost and computational cost should be measured and controlled to an acceptable level.

In order to achieve spatial-temporal knowledge transfer in FCIL under a trustworthy framework, we propose FedProK (\textbf{Fed}erated \textbf{Pro}totypical Feature \textbf{K}nowledge Transfer). FedProK consists of two components: (1) \textbf{feature translation} procedure on the client side that achieves temporal knowledge transfer among tasks to mitigate catastrophic forgetting, and (2) \textbf{prototypical knowledge fusion} on the server side that enables spatial knowledge transfer across clients to address data heterogeneity.

The contributions of this paper are as follows:
\begin{enumerate}
    \item We raise the issue of trustworthiness in FCIL and formulate a comprehensive framework of trustworthy FCIL concerning continual utility, efficiency, and privacy.
    \item We propose FedProK to achieve temporal knowledge transfer among tasks and spatial knowledge transfer across clients under the trustworthy FCIL framework.
    \item We evaluate the effectiveness of our method in knowledge transfer with extensive experiments, including various settings of class increments and data heterogeneity.
\end{enumerate}

\section{Related Work}
\subsection{Federated Class-Incremental Learning}
Class Incremental Learning (CIL) methods aim to overcome catastrophic forgetting of the previous classes when training on the new classes \cite{zhao2020maintaining,kirkpatrick2017overcoming,rebuffi2017icarl,petit2023fetril}. With the growing awareness of data privacy, FCIL has emerged as a novel approach of dynamic, heterogeneous and distributed machine learning paradigm \cite{yang2023federated}. 

FCIL can be categorized into synchronous and asynchronous based on whether the task orders are the same among clients \cite{yang2023federated}. Early efforts often directly integrate existing CL methods in FL. FedCurv \cite{shoham2019overcoming} adds a regularization term in the loss function to penalize the changes in important parameters of previous tasks. GradMA\cite{luo2023gradma} utilizes quadratic programming to correct the direction of local updates. \cite{ma2022continual} performs knowledge distillation on both the client and the server side according to a surrogate dataset. Recently, studies have started to focus on more complex scenarios and methods. FedSW \cite{zhao2024fedsw} jointly optimize the global model and local models with self-paced adaptive sample weights. FedSSD \cite{he2022learning} transfers the knowledge in the previous global model to the local updates in the current round with a selective knowledge distillation mechanism. Besides knowledge distillation and regularization, experience replay is also commonly employed in FCIL \cite{yang2023federated}. FedWeIT \cite{dong2022federated} constructs an exemplar memory to retain the samples from the previous classes and performs rehearsal when training on new tasks. TARGET \cite{zhang2023target} replays globally synthesized pseudo data on the clients to mitigate forgetting and alleviate different degrees of heterogeneity.

Generally, current FCIL methods predominantly focus on overcoming catastrophic forgetting, ignoring efficiency and privacy. Some methods even store and replay a shared core set of previous classes \cite{wei2022knowledge,zhang2023target}, severely compromising the privacy of FL. Besides, the exploration of stability and plasticity dilemma in FCIL remains unaddressed \cite{yoon2021federated,criado2022non}. 

\subsection{Trustworthy Federated Learning}
FL is naturally more aligned with the core of trustworthy AI than the centralized training paradigm \cite{bagwe2023fed,kang2020reliable}. To enhance its trustworthiness, researchers delve into the privacy, efficiency, fairness, and interpretability of FL. \cite{yang2022trustworthy} designs a blockchain-based FL framework to ensure systematic security and privacy. \cite{lo2022toward}
focuses on fairness and accountability of blockchain-based FL to improve the trustworthiness of medical diagnoses. \cite{zhu2019multi} adopts the elitist non-dominated sorting genetic algorithm (NSGA-II) to minimize the communication overhead and maximize the model accuracy. \cite{jin2022evolutionary} utilizes scalable encoding methods and real-time federated evolutionary strategies to optimize neural network models for reducing communication costs and improving performance on edge devices.

A trustworthy FL system typically involves more than one objective \cite{kang2022framework,gu2021novel} to optimize. \cite{zhang2023tradeoff} demonstrates that simultaneously optimizing utility, privacy, and efficiency to its optimal is impossible. For instance, differential privacy adds Gaussian or Laplace noise to obfuscate the network parameters transmitting between clients and the server, but excessive noise can severely degrade the accuracy of the model \cite{zhang2022no}. Similarly, while homomorphic encryption safeguards data privacy by translating parameters into ciphertext, overly sophisticated encryption and decryption processes may reduce model efficiency \cite{zhang2022trading}. 

Current works on trustworthy FL mainly focus on stationary data and there is no trustworthy FCIL framework yet \cite{lin2020ensemble,arivazhagan2019federated}. However, in the real-world scenario, a dynamic and heterogeneous FCIL setting is worth investigating.
\section{Problem Definition}
\subsection{Federated Class-Incremental Learning}

As the combination of FL and CIL, there is a class-incremental task sequence $\mathcal{T}_k=\{\mathcal{T}^1_k,\dots,\mathcal{T}^T_k\}$ on each local client $k$ in FCIL, where $T$ is the total number of incremental states. The composition of classes in the local task sequences can either be the same across clients (i.e., synchronous FCIL) or different across clients (i.e., asynchronous FCIL). Not every communication round $r \in R$ involves class increments \cite{de2021continual}. Communication rounds alternate between incremental state and interval rounds, and the dataset $\mathcal{D}^t$ and its class set $\mathcal{C}^t$ of the interval rounds are consistent with those of the latest incremental state. We denote the local task at $t$-th increments and its subsequent interval rounds on client $k$ as $\mathcal{T}^t_k=\{(\textbf{x}_i,\textbf{y}_i)\}^{N^t}_{i=1}$, where $N^t$ denotes the number of samples of task $t$. Thus, the entire label space at $t$-th task is $\mathcal{Y}^t=\cup ^K_{k=1}\mathcal{Y}^t_k$, which includes:
\begin{itemize}
    \item New classes: $\mathcal{C}^t=\cup^K_{k=1}\mathcal{C}^t_k$,
    \item Previous classes: $\cup^{t-1}_{i=1}\mathcal{C}^i$.
\end{itemize}

The standard objective of FCIL is to train a global model $\Theta _G$ that is capable of discriminating samples within the entire label space $\mathcal{Y}^T$, i.e., the superset of classes observed by all of the clients in the federated system. It is mainly challenged by catastrophic forgetting and data heterogeneity issues. Catastrophic forgetting refers to the severe degradation of model performance in previous classes. Data heterogeneity among clients incurs the deviation between the individual optima of the local models and the true optimum of the global model \cite{li2020federated}. In FCIL, particularly, data heterogeneity exacerbates catastrophic forgetting by overwriting the important weights of previous classes more severely \cite{yang2023federated}. In this paper, we explore the intertwined challenge of class-incremental learning among tasks and data heterogeneity among clients. 

\subsection{Synchronous and Asynchronous Federated Class-Incremental Learning}
FCIL can be categorized into synchronous FCIL and asynchronous FCIL according to the different forms of data heterogeneity \cite{yang2023federated}. 

\textbf{Synchronous FCIL.} In this scenario, the task $\mathcal{T}_k$ at the same incremental state $t$ contains the data of identical classes $\mathcal{C}^t_k=\mathcal{C}^t_K$ but varied amounts of samples for each class. It is caused by distribution-based label imbalance among clients \cite{li2020federated}. Each client is allocated a proportion of the samples of each label according to Dirichlet distribution. The degree of heterogeneity is controlled with the concentration parameter of Dirichlet distribution $\alpha$. A smaller value of $\alpha$ results in a more heterogeneous data partition. 

\textbf{Asynchronous FCIL.} In this scenario, the task $\mathcal{T}_k$ at the same incremental state $t$ contains the data of different classes $\mathcal{C}_k^t \neq \mathcal{C}_l^t$ ($k,l\in K$). It is caused by quantity-based label imbalance among clients \cite{li2020federated}. We refer to the proportion of overlapping classes in the total number of classes $\cup ^T_{t=1}\mathcal{C}^t_k \cap \cup ^T_{t=1}\mathcal{C}^t_l$ $(k,l\in K)$ in the local task sequence $\mathcal{T}_k, \mathcal{T}_l$ as consensus rate $\gamma$. For instance, if each client's local task sequence comprises samples of 50 classes, with 25 common classes and 25 unique classes, the consensus rate $\gamma$ in this case is 0.5. Intuitively, a lower $\gamma$ in the federated system indicates a higher degree of heterogeneity. 

Notably, in both cases, we followed the previous works \cite{yoon2021federated,dong2022federated,wang2023federated} to assume that the numbers of clients $K$ are consistent during the entire training.

\section{Trustworthy Framework of Federated Class-Incremental Learning}
\label{frame}
By achieving spatial-temporal knowledge transfer among tasks and across clients, the FCIL global model gains the capability of discriminating the superset of classes observed by all clients in the federated system. There has been plenty of work dedicated to designing such a model. However, existing works exhibit limitations in the following aspects:
\begin{itemize}
    \item Overemphasizing the preservation of previous knowledge (stability) while ignoring the accommodation of new knowledge (plasticity).
    \item A complicated model design may increase the computational cost of local update and aggregation, and a massive number of parameters transmitted between clients and the server may increase the communication cost.
    \item Some approaches to mitigate forgetting and data heterogeneity may break the privacy protocol of FL, especially exemplar-based approaches.
\end{itemize}

A trustworthy FL algorithm must satisfy multiple conflicting objectives simultaneously, which includes three major objectives, i.e., privacy, efficiency, and utility, and several minor objectives, i.e., robustness, interpretability, fairness, etc. \cite{kang2023optimizing,hu2022federated}. Therefore, we propose a trustworthy framework of FCIL to consolidate multiple objectives:
\begin{itemize}
    \item \textbf{Continual Utility}: trustworthy FCIL models must handle the classic \textbf{plasticity-stability} dilemma inherent in CL with more effective spatial-temporal knowledge transfer. 
    \item \textbf{Privacy}: as a main driving force of FL, \textbf{data privacy} should not be neglected. In this way, a shared buffer of private data must be avoided.
    \item \textbf{Efficiency}: mitigating forgetting requires extra training costs, like knowledge distillation and generative replay. \textbf{Computational cost and communication cost} should be measured and controlled to an acceptable range.
\end{itemize}

Utility refers to the model's performance, typically assessed through the average accuracy \cite{chai2023survey,yang2019federated}, which is not sufficient for FCIL. To find trade-offs between the stability-plasticity dilemma of continual learning, we evaluate stability and plasticity, respectively, with averaged accuracy on the test set of previous classes and current classes during the entire training. After the completion of training on task $t$, the test accuracy on the previous tasks $\mathcal{A}^{t-1}_G$ and on the current task $\mathcal{A}^{t}_G$ reflect the stability and plasticity of the global model $\theta_G$. Note that the test set to produce $\mathcal{A}^{t-1}_G$ contains samples of  $\cup^{t-1}_{i=1}\mathcal{C}^i$ (the classes of previous tasks) and the test set to produce $\mathcal{A}^{t}_G$ contains samples of $\mathcal{C}^t$ (the classes of the current task). Based on this, we formulate the following equation to assess continual utility, i.e. the trade-off between stability and plasticity in FCIL: 
\begin{equation}
\label{utility}
    U(\theta_G)=\lambda \times \mathcal{A}^{t-1}_G+(1-\lambda) \times \mathcal{A}^{t}_G,
\end{equation}
where $\lambda$ is the hyperparameter and $\theta_G$ is the parameters of the global model $G$.

Privacy refers to the effectiveness of avoiding the leakage of the original information \cite{dong2022federated}. We assess privacy by measuring the similarity between the reconstructed images generated by malicious attackers and the original local images of clients, which is defined as follows:
\begin{equation}
\label{Privacy}
    P(\theta_G)=1-\frac{1}{1+\text{MSE}(img_{true},img_{pred})},
\end{equation}
where $\text{MSE}$ is the mean square error, $img_{true}$ and $img_{pred}$ denote the numpy arrays of the ground truth image and the reconstructed image.

Efficiency refers to the model's total costs of communication and computation. We assess the efficiency of methods with the time to perform one epoch of federated iteration \cite{wang2023federated}, which is defined as:
\begin{equation}
    E(\theta_G)=\frac{1}{R}(\tau_1+\tau_2),
\end{equation}
where $R$ denotes the number of rounds, $\tau_1$ and $\tau_2$ denote the communication cost and computation cost. $\tau_1$ reflects the overhead of data transmission between server and clients, and $\tau_2$ reflects the total overhead of local updates and model aggregation.

\section{The Proposed Method}
\begin{figure*}[htbp]
  \centering
  \includegraphics[width=0.98\textwidth]{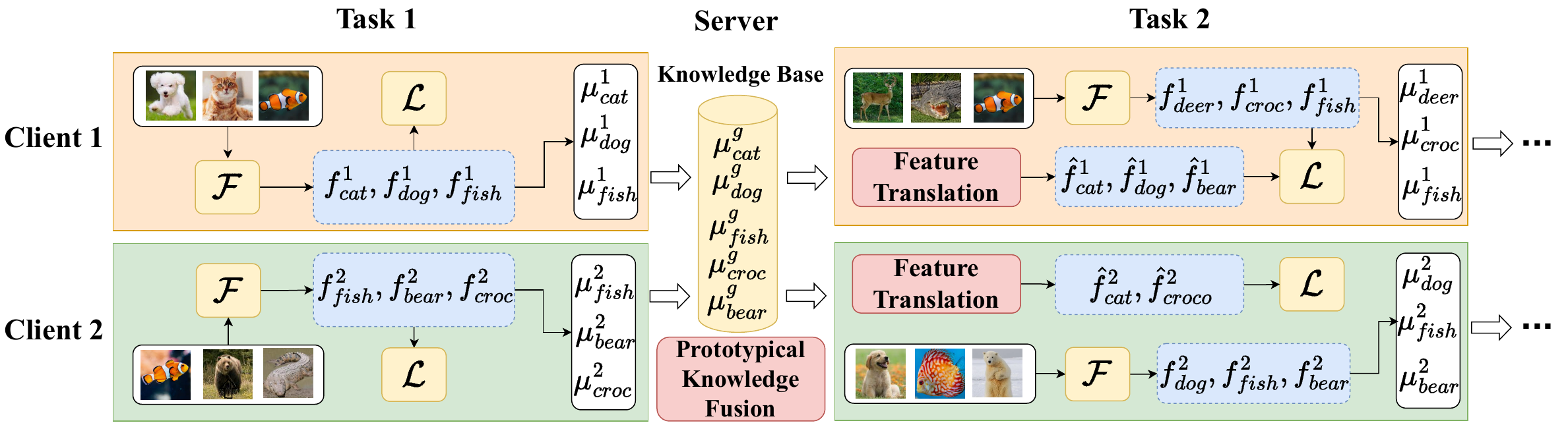}
  \caption{Illustration of the knowledge transfer via FedProK. The clients conduct feature translation locally to mitigate forgetting of the previous classes of the entire federated system. The server conducts prototypical knowledge fusion to alleviate the adverse impact of data heterogeneity among clients. $\mathcal{F}$ and $\mathcal{L}$ denote the feature extractor and the classification layer.}
  \label{fig_method}
\end{figure*}

\begin{algorithm}[tb]

\caption{FedProK}
\label{alg:algorithm}
\textbf{Input}: local dataset $\{D_1,\dots,D_K\}$, global model $G$ with initial weights $\theta^0_G$, number of local epochs $E$, number of communication rounds $R$, number of incremental tasks $T$\\
\textbf{Output}: global model $G$ with final weights $\theta^R_G$ \\
\begin{algorithmic}[1] 
\FOR{$r=1$ to $R$}
\FOR{each client $k$}
\STATE Compute task number $t=r//(R/T)+1$
\STATE Compute prototype $\mu^{t}_{k,c}$ for each new class $c$
\IF{$t>1$}
\STATE Compute pair-wise relations of previous classes and new classes, select the base class $n \in \mathcal{C}^t$ of feature translation for each previous class $p \in \cup^{t-1}_{i=1}\mathcal{C}^i$: $n=\mathop{\arg\min}\limits_n\text{dis}(\mu^t_{k,p},\mu^{t}_{k,c})$
\STATE Perform feature translation for each previous class $\hat{f}^t(\textbf{x}_p)=f(\textbf{x}_n)+\mu ^t_{k,p}-\mu ^t_{k,n}$
\ENDIF
\STATE Local training on current task $\mathcal{T}_k^t$ with pseudo feature $\hat{f}^t(\textbf{x}_p)$
\STATE Upload local weights $\theta^r_k$
\STATE Transfer local prototypical knowledge $\mathcal{P} ^r_k$\\
\ENDFOR
\STATE Aggregate model weights with standard FedAvg $\frac{1}{K}\sum^K_{k=1}\theta^r_k$
\STATE Construct the prototypical knowledge base $\mathcal{P}^t_G$
\STATE Distribute weights of global model $\theta^r_G$ to clients
\STATE Transfer global prototypical knowledge $\mathcal{L}^r_G$ to clients\\
\ENDFOR
\end{algorithmic}
\end{algorithm}

The overview of FedProK is depicted in \cref{fig_method} and summarized in \cref{alg:algorithm}. FedProK comprises two novel components: a client-side feature translation procedure to achieve temporal knowledge transfer and a server-side prototypical knowledge fusion mechanism to achieve spatial knowledge transfer. During the local update phase, clients train on their private data, compute the prototype for the new classes, and perform feature translation to mitigate forgetting of the previous classes. The prototype lists, as well as the weights of local networks, are uploaded to the server. During the global aggregation phase, the server needs to perform a standard FedAvg to aggregate the model parameters and a prototypical aggregation to construct a global knowledge base. Then, the class-wise prototype list is distributed back to the clients. Since the knowledge base contains the global prototypical list of previous classes without reserving or globally generating any exemplars, FedProK can effectively transfer knowledge across clients and among tasks without compromising data privacy.

\subsection{Feature Translation}
To alleviate the catastrophic forgetting of past knowledge, we utilize a feature translation mechanism locally to enable temporal knowledge transfer among tasks. Feature translation module is a combination of a real feature extractor $\mathcal{F}$ and a pseudo-feature generator $\mathcal{G}$. Firstly, $\mathcal{F}$ extracts the feature vectors of sample $(\textbf{x}_i,\textbf{y}_i)$:
\begin{equation}
    f(\textbf{x}_i)=\mathcal{F}(\textbf{x}_i,\textbf{y}_i).
\end{equation}
After that, client $k$ computes the local prototype $\mu$ for each class and reserves it in the local prototypical list $\mathcal{L}^t_k$:
\begin{equation}
    \mu_{k,c}^t=\frac{1}{\left|D_{k,c}\right|}\sum^{\left|D_{k,c}\right|}_{i=1}f(\textbf{x}_i),
\end{equation}
where $\left|D_{k,c}\right|$ denotes the number of local samples belonging to class $c$ on client $k$. When new data arrives, $\mathcal{F}$ is frozen, and the generator $\mathcal{G}$ generates pseudo feature vectors of previous classes according to the real feature of a selected new class and the prototype of the past classes. 

Generally, each incremental state may involve more than one new class, there are several strategies for selecting the base class of feature translation. The simplest strategy is to randomly select one from all new classes as the base class. However, we observed that the random selection performs poorly when the inter-class relation is relatively large. Therefore, for a previous class $p \in \cup^{t-1}_{i=1}\mathcal{C}^i$ and a new class $c \in \mathcal{C}^t$,  we compute the pair-wise relation with cosine similarity:
\begin{equation}
    \text{dis}(\mu^t_{k,p},\mu^{t}_{k,c})=\frac{\mu^t_{k,p}\cdot\mu^{t}_{k,c}}{\left \|\mu^t_{k,p}\right \| _2\times \left \|\mu^{t}_{k,c}  \right \| _2}.
\end{equation}

And we select the nearest new class $n$ as the geometric base of feature translation for class $p$ to simulate the representation of the previous class $p$ to the greatest extent: 
\begin{equation}
     n=\mathop{\arg\min}\limits_c\text{dis}(\mu^t_{k,p},\mu^{t}_{k,c}).
\end{equation}
Through the translation process in \cref{eq1}, the pseudo feature vector is generated by translating the value of each dimension with the distance between the values of the corresponding dimension of $\mu ^t_{k,n}$ and $\mu ^t_{k,p}$. Rehearse them by directly sending them to the classification layer alongside new inputs.

Since $\mathcal{F}$ is all layers of the backbone except for the classification layer, the straightforward feature translation process is much more computationally efficient than other generative replay methods \cite{petit2023fetril}, i.e., Generative Adversarial Net (GAN) and Variational Auto-Encoder (VAE). At task $t$, the feature translation from a certain new class $n \in \mathcal{C}^t$ to previous class $p\cup^{t-1}_{i=1}\mathcal{C}^i$ is defined as: 
\begin{equation}
    \hat{f}^t(\textbf{x}_p)=f(\textbf{x}_n)+\mu ^t_{k,p}-\mu ^t_{k,n},
    \label{eq1}
\end{equation}
where $f(\textbf{x}_n)$ denotes the feature of a real sample $\textbf{x}_n$ in class $n$ extracted by $\mathcal{F}$, and $\hat{f}^t(\textbf{x}_p)$ denotes the pseudo feature vector of a pseudo sample $\textbf{x}_p$ of class $p$ produced in the $t$-th incremental state by $\mathcal{G}$. Notably, $\mu ^t_{k,n}$ is currently computed while $\mu ^t_{k,p}$ is previously stored in the $\mathcal{P}^t_k$.
\subsection{Prototypical Knowledge Fusion}
To address data heterogeneity, we construct a knowledge base on the server with prototypical knowledge fusion mechanism to achieve spatial knowledge transfer among clients. At each round $r$ with its corresponding task $t$, client $k$ computes the local prototype for each class $c \in \mathcal{C}^t_k$ and uploads the local prototypical list $\mathcal{P}_k$ to the server. 

In synchronous FCIL and asynchronous FCIL with a consensus rate $\gamma$ greater than zero, local class sets overlap across clients, which is defined as $\mathcal{C}^t_i \cap \mathcal{C}^t_j \neq \emptyset (i,j \in K$). In this case, the uploaded local prototypical lists $\mathcal{P}_i$ and $\mathcal{P}_j$ might contain different prototypes $\mu^t_{i,c} \neq \mu^t_{j,c}$ for the same class $c$. Besides, if the current dataset includes previous classes, the knowledge base $\mathcal{P}^t_G$ on the server should be updated to rectify the bias of previous data and alleviate the impacts of concept drift \cite{de2021continual}. Therefore, we design the prototypical knowledge fusion mechanism that horizontally aggregates heterogeneous prototypical knowledge from different clients and vertically fuses previous and new knowledge along the timeline. The fusion process is defined as:
\begin{equation}
    \begin{split}
\mu _c^{t}= \left \{
\begin{array}{lr}
    \sum^K_{k=1} \frac{|D_{k,c}|}{|D_c|}\mu ^{r}_{k,c} ,  &c\notin \cup^{t-1}_{i=1}\mathcal{C}^i\\
    \\
    \beta (\sum^K_{k=1} \frac{|D_{k,c}|}{|D_c|}\mu ^{r}_{k,c})+(1-\beta)\mu_c^{t-1},     & otherwise\\
\end{array}
\right.
\end{split}.
\end{equation}
Then, the server reserves the fused knowledge $\mathcal{P}^t_G$ in the knowledge base and distributes it to the participants of the next communication round. $\mathcal{P}^t_k$ is overwritten by $\mathcal{P}^t_G$ before the local training to realize the spatial knowledge transfer.

\section{Experiments}
\begin{table*}[h]
    \centering
    \renewcommand\arraystretch{1.2}
    \begin{tabular}{c|cc|cc|cc|cc|cc}
    \hline
        Setting&\multicolumn{8}{c|}{Synchronous FCIL}&\multicolumn{2}{c}{Asynchronous FCIL}\\
        \hline
        Dataset&\multicolumn{4}{c|}{CIFAR-10}&\multicolumn{6}{c}{CIFAR-100}\\
        \hline
        \multirow{2}{*}{Method}& \multicolumn{2}{c|}{T=2}  & \multicolumn{2}{c|}{T=5} & \multicolumn{2}{c|}{T=5}& \multicolumn{2}{c|}{T=10}& \multicolumn{2}{c}{T=5}\\
        
          & $\alpha=1$ & $\alpha=0.5$ & $\alpha=1$ & $\alpha=0.5$&$\alpha=1$ & $\alpha=0.5$ & $\alpha=1$ & $\alpha=0.5$&$\gamma=0.5$&$\gamma=0.2$\\
          \hline
        FedAvg & 43.59 & 41.17 & 18.45 & 16.53 & 16.42& 15.92 &  8.66& 8.40&2.78& 1.03\\
        FedEWC& 43.13 &  40.05& 17.92 & 15.22& 16.86 &  16.06& 8.84 & 8.04&3.57& 1.02\\
        FediCaRL & 60.01 & 58.14 & 26.81 & 25.17& 24.25& 20.51 & 18.50 & 14.14&9.66& 2.28\\
        FedProx & 46.27 & 43.55 & 17.49& 17.05 & 19.43& 19.59 &  10.82& 9.12&3.23& 1.97\\
        FedNova &  46.20& 44.17 & 19.61 & 18.14&18.97 &19.88  &  11.01&8.96 &4.13& 1.15\\
        GLFC & 72.34&  67.99 &  55.80& 52.74& \underline{36.88}&31.43  & \underline{26.19} &\underline{24.54} &13.59& 9.92\\
        FedSpace & 55.15 & 50.60 & 21.60 &19.56 &21.23 & 21.35 &  19.00& 18.24&4.15& 3.76\\
        TARGET & 76.52 & 74.91 & \underline{59.66}&  \underline{57.37}&  34.89& \underline{33.33}& 22.85 &20.71 &\underline{14.17}& \underline{11.67}\\
        \hline
        Ours-w/oFT&43.74&41.10&17.73&16.25&16.55&15.73&8.76&8.28&-&-\\
        Ours-w/oPKF&\underline{78.34}&\underline{76.73}&58.76&56.27&35.23&34.88&24.90&24.17&-&-\\
        Ours (FedProK)&\textbf{87.62}&\textbf{84.13}&\textbf{64.34}&\textbf{62.79}&\textbf{41.02}&\textbf{39.61}&\textbf{26.94}&\textbf{26.07}&\textbf{18.52}& \textbf{12.21}\\
    \hline
    \end{tabular}
    \caption{Final accuracy (\%) of the global model on CIFAR-10 and CIFAR-100, where $\alpha$ denotes the Dirichlet concentration parameter in synchronous setting and $\gamma$ denotes the consensus rate in asynchronous setting. The degree of heterogeneity increases from the left to the right end of the table. }
    \label{tab_accuracy}
\end{table*}

In this section, we consider a realistic FCIL training scenario composed of an alternate local class-incremental update and global federated iteration. We conduct comprehensive experiments under synchronous and asynchronous FCIL settings to evaluate FedProK in terms of accuracy and trustworthiness. 
\subsection{Experiment Setting}

\textbf{Datasets.} We conduct experiments on two datasets: (1) CIFAR-10 \cite{krizhevsky2014cifar}: 10 classes, 32$\times$32 pixels images, 5000 training and 1000 test images per class. (2) CIFAR-100 \cite{krizhevsky2014cifar}: 100 classes, 32$\times$32 pixels images, 500 training and 100 test images per class. 

\textbf{Synchronous FCIL Setting.} In the synchronous setting, clients share the same class-incremental task sequence but suffer a distribution-based label imbalance, which is controlled with the Dirichlet concentration parameter $\alpha$. We partitioned the whole dataset into $T$ tasks and apportioned them to $N$ clients according to $\alpha$. For CIFAR-10, we set $N=3, T=2$ and $N=3,T=5$, each task contains data of $5$ or $2$ classes and $\alpha=\{0.5,1\}$. For CIFAR-100, we set $N=3, T=5$, $N=3, T=10$, each task contains data of $20$ or $10$ classes and $\alpha=\{0.5,1\}$. Specifically, not every communication round involves the arrival of new tasks. We set the interval of incremental states to be every $5$ round for CIFAR-10 and every $10$ round for CIFAR-100. 

\textbf{Asynchronous FCIL Setting.} In the asynchronous setting, some of the classes are accessible to all clients, while others are private to a single client. We refer to the rate of common classes as \textbf{consensus rate}, denoted with $\gamma$. Intuitively, the smaller $\gamma$ indicates a higher degree of heterogeneity. Due to the limited number of classes in CIFAR-10, we only conduct asynchronous FCIL experiments on CIFAR-100. We set $N=3, T=5, \gamma=0.2$ and $N=3, T=5, \gamma=0.5$, which means $10$ or $25$ classes are common on all clients while the rest of $40$ or $25$ classes are unique on one client. We set the interval of incremental states to be every $10$ round under this setting.

\textbf{Baselines.} We compare FedProK with static FL baselines, existing FCIL methods, and hybrid methods, i.e., combination of FL and CL methods. Static FL methods include FedAvg \cite{mcmahan2017communication}, FedProx \cite{li2020federated} and FedNova \cite{wang2020tackling}. FCIL competitors include GLFC \cite{dong2022federated}, FedSpace \cite{shenaj2023asynchronous} and TARGET \cite{zhang2023target}. Hybrid competitors include FedEWC \cite{kirkpatrick2017overcoming} and FediCaRL \cite{rebuffi2017icarl}. More detailed descriptions are in the Appendix.

\textbf{Evaluation Metric.} We utilize the final accuracy of the global model, denoted as $\mathcal{A}^R_G$, as the metric to showcase the performance of our model and baselines. This choice aligns with the primary goal of FCIL, which focuses on training a global model capable of distinguishing between all classes. The trustworthiness is assessed with $U(\theta_G)$, $P(\theta_G)$ and $E(\theta_G)$. Definitions of the metrics are in \cref{frame}.

\textbf{Implementation Details.} The code of our method and baselines is implemented with PyTorch. The backbone network is ResNet-18. We conducted experiments on each dataset using three random seeds (42, 1999, 2024) and averaged the results. The size of the exemplar set for all of the exemplar-based methods is $500$. All the experiments were conducted sequentially on NVIDIA RTX-3090 GPU. 

\subsection{Experimental Results}

\subsubsection{Results on Accuracy}
\begin{figure}[t]
    \centering
    \includegraphics[width=0.48\textwidth]{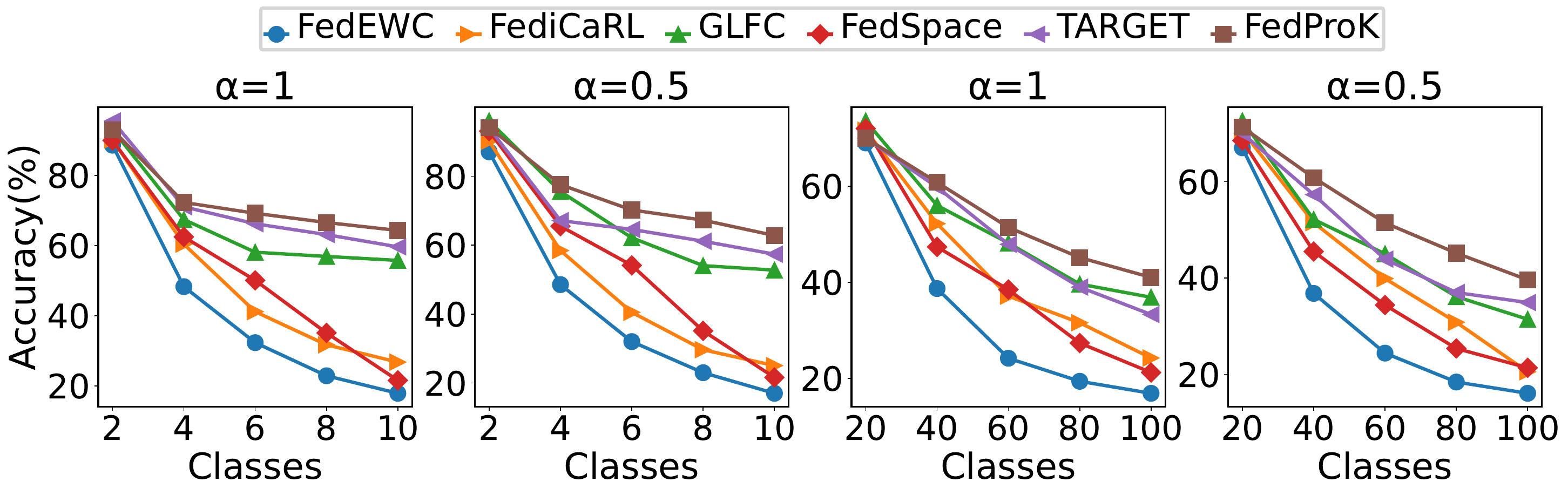}
    \caption{Experimental results on CIFAR-10 (left) and CIFAR-100 (right) with 5 incremental tasks, where $\alpha$ denotes the Dirichlet concentration parameter.}
    \label{fig_task5}
\end{figure}

\begin{figure}[t]
    \centering
    \includegraphics[width=0.48\textwidth]{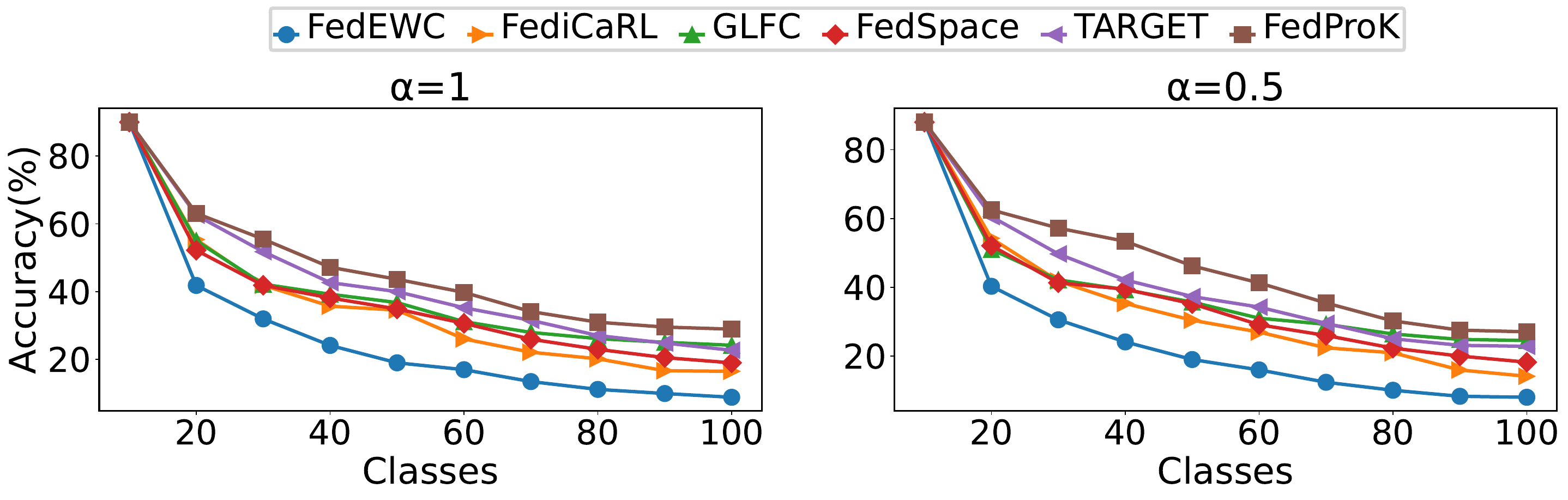}
    \caption{Experimental results on CIFAR-100 with 10 incremental tasks, where $\alpha$ denotes the Dirichlet concentration parameter.}
    \label{fig_task10}
\end{figure}

\cref{fig_task5} and \cref{fig_task10} illustrate the test accuracy of the global model on all learned tasks after training on the current task. As shown in \cref{fig_task5} (left), although CIFAR-10 has a limited number of total classes, the global models still suffer significant performance decline with the arrival of new classes. Despite the initial round, FedProK consistently outperforms other FCIL methods by the margin of $9.22\%\sim44.08\%$ when $\alpha=1$ and $5.42\%\sim47.57\%$ when $\alpha=0.5$. In \cref{fig_task5} (right), FedEWC completely failed in preventing catastrophic forgetting of the previous tasks, while FedSpace and FediCaRL are just slightly better. Our method outperforms the competitive GLFC and TARGET by $4.14\%\sim7.72\%$ when $\alpha=1$ and $4.72\%\sim8.18\%$ when $\alpha=0.5$. \cref{fig_task10} illustrates the results on CIFAR-100 with 10 incremental tasks when $\alpha=1$ and $\alpha=0.5$, in which our FedProK maintains the highest test accuracy during the entire training.

The final accuracy of the global model on CIFAR-10 and CIFAR-100 are reported in \cref{tab_accuracy}. The level of data heterogeneity among clients gradually increases from the left to the right end of the table, and the performance of each method correspondingly decreases. Static FL methods (FedAvg, FedProx, and FedNova) failed in both synchronous FCIL and asynchronous FCIL settings, indicating that the dynamically incremented classes adversely impact the model performance. The regularization-based hybrid method (FedEWC) showed even poorer performance than heterogeneous FL methods (FedProx and FedNova), implying that EWC is incapable of FCIL. We observed that our model outperformed existing FCIL methods (GLFC, FedSpace, and TARGET) under all settings. FedProK also achieved better accuracy by $8.54\%\sim0.54\%$ under the most challenging scenario, i.e., asynchronous FCIL with $\gamma=0.2$. The experimental results highlight the effectiveness of our method in overcoming catastrophic forgetting and data heterogeneity. 
\subsubsection{Results on Trustworthiness}
We compared the trustworthiness of our method with baselines in terms of continual utility, efficiency, and privacy. As stated in \cref{frame}, we validate the continual utility of our method and baselines with the trade-off between stability and plasticity, which are measured with the global accuracy of the previous classes and current classes respectively. Efficiency is measured with the average training time per communication round, reflecting the total cost of communication and computation of federated training. Privacy is measured by the similarity between the reconstructed images generated by malicious attackers and the original images, where a larger result indicates better performance in privacy preservation. 

\begin{table}
    \centering
    \renewcommand\arraystretch{1.2}
    \begin{tabular}{c|ccc}
        \hline
        \multirow{2}{*}{Method} & Continual  & Efficiency& Privacy\\
        & Utility(\%)&(s)&($10^{-3}$)\\
        \hline
        FedEWC & 21.69 & \textbf{79.72} & 2.4763\\
        FediCaRL & 31.37 & 94.27 & 0.5569 \\
        GLFC & 45.04 & 165.33 & 11.7101\\
        TARGET & 45.77 & 142.10& -\\
        \hline
        Ours (FedProK) & \textbf{53.23} & 118.04& \textbf{25.6118} \\
        \hline
    \end{tabular}
    \caption{Continual utility, efficiency, and privacy of our method and baselines on CIFAR-100 with 5 class-incremental states, which are measured with \cref{utility} ($\lambda=0.5$), training time per epoch, and whether the FCIL method is exemplar-based.}
    \label{trust}
\end{table}

Based on the experimental results reported in \cref{trust}, FedProK exhibited superior trustworthiness compared with various baselines. Specifically, our method outperformed baselines on continual utility by $7.46\% \sim 31.54\%$ when $\lambda=0.5$. And FedProK showed better efficiency than GLFC and TARGET, with $47.29$ and $24.06$ seconds per communication round, respectively. Besides, FedProK also outperformed in data privacy by $25.0549\sim13.9017$.

\begin{figure}[t]
  \centering
  \includegraphics[width=0.48\textwidth,height=0.32\textwidth]{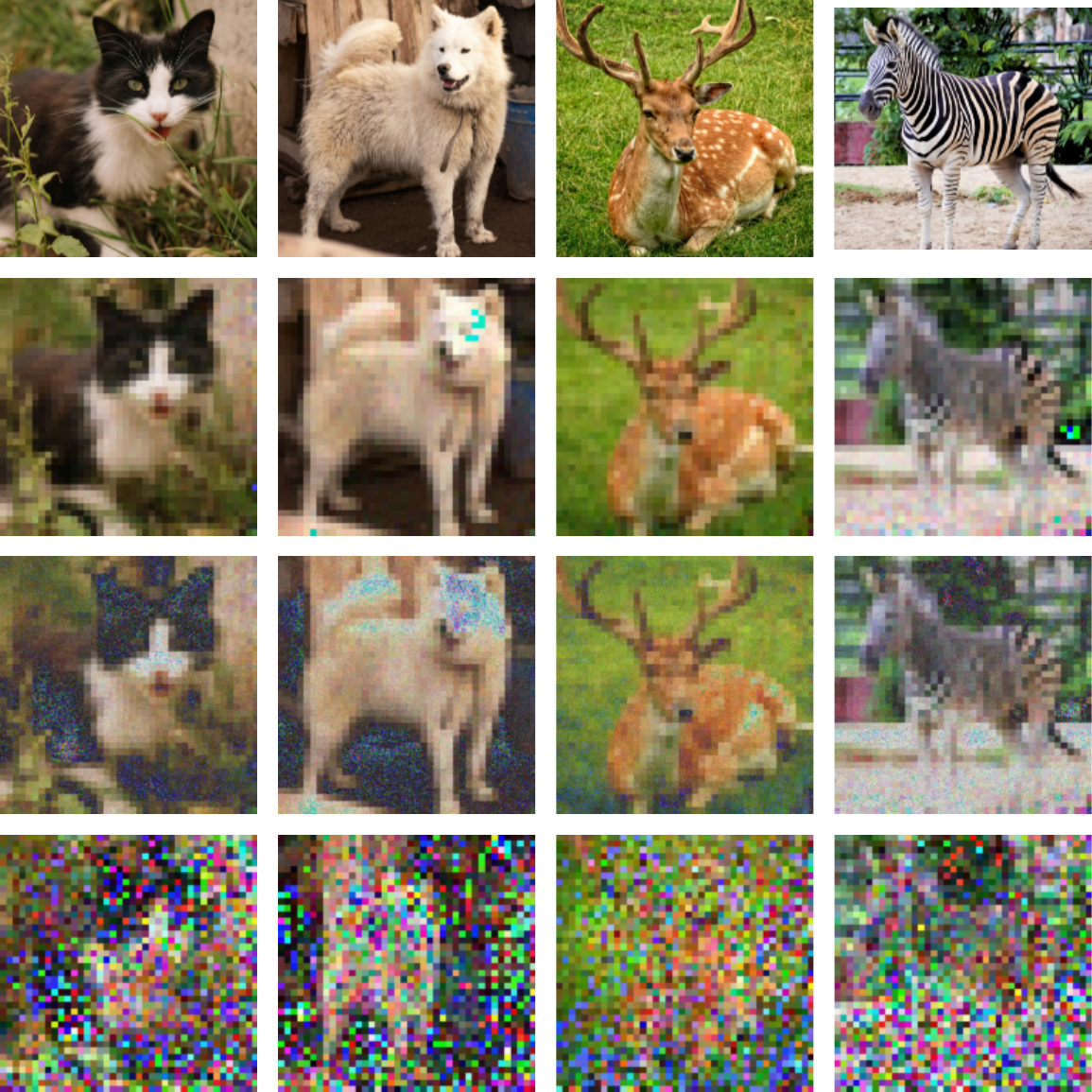}
  \caption{Illustration of the reconstructed images of malicious attackers, where the first line is the ground truth, the second line is the reconstructed images of FediCaRL, the third line is the reconstructed images of GLFC and the bottom line is the reconstructed images of our method.}
  \label{fig_privacy}
\end{figure}

\cref{fig_privacy} illustrates the reconstructed images of malicious attackers \cite{zhu2019deep}. As an exemplar-based method, FediCaRL suffers the most significant data leakage where the private images are completely reconstructed by attackers. GLFC exhibits some robustness against attacks because it adds Gaussian noise to the stored exemplars \cite{dong2022federated}. The reconstructed images of FedProK show the greatest differences with the ground-truth images, affirming the superiority of our exemplar-free approach to privacy preservation.

\begin{figure}
  \centering
  \begin{subfigure}{0.51\linewidth}
    \includegraphics[width=\linewidth]{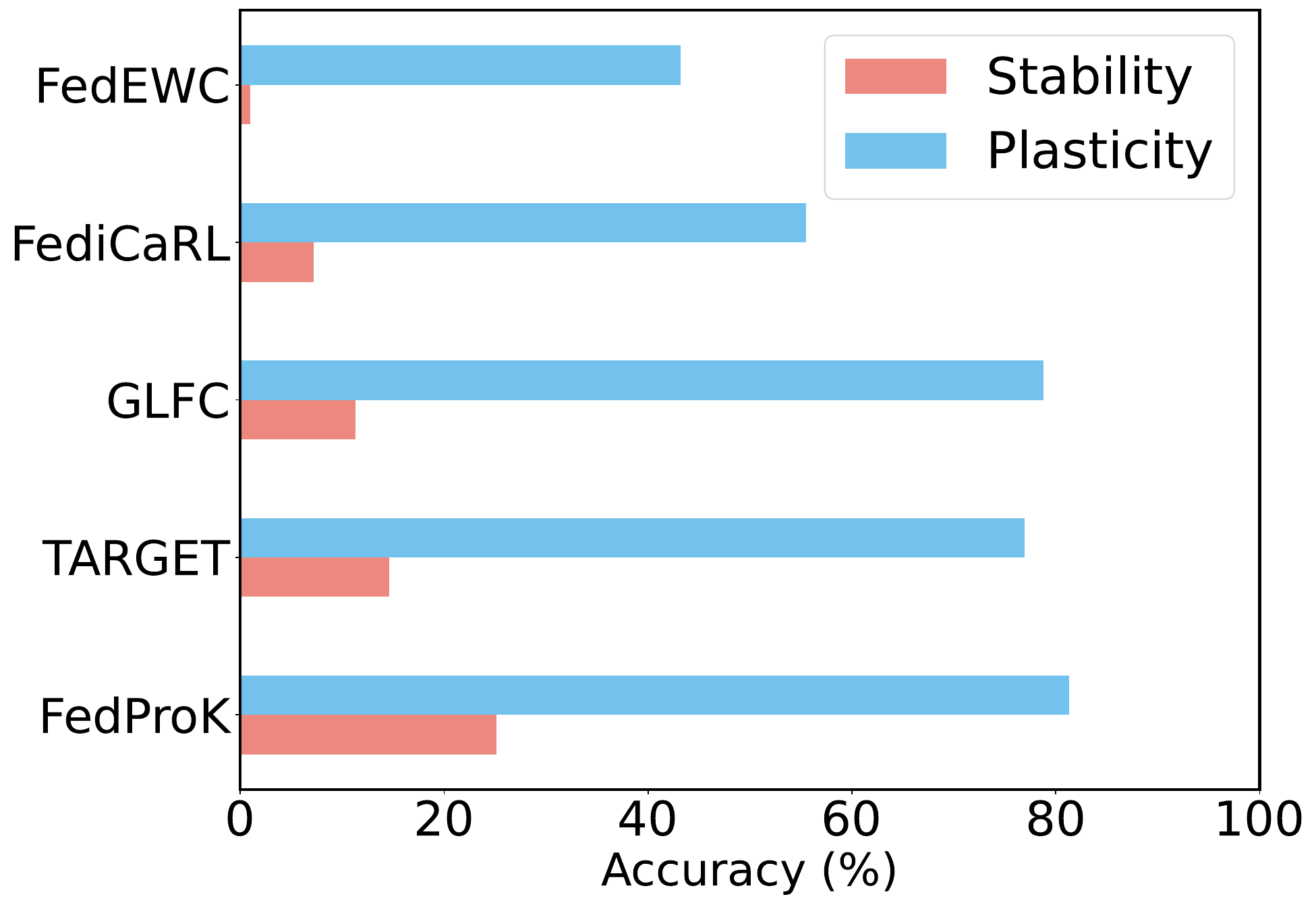}
    \caption{Stability and plasticity.}
    \label{tradeoff}
  \end{subfigure}
  \hfill
  \begin{subfigure}{0.40\linewidth}
    \includegraphics[width=\linewidth]{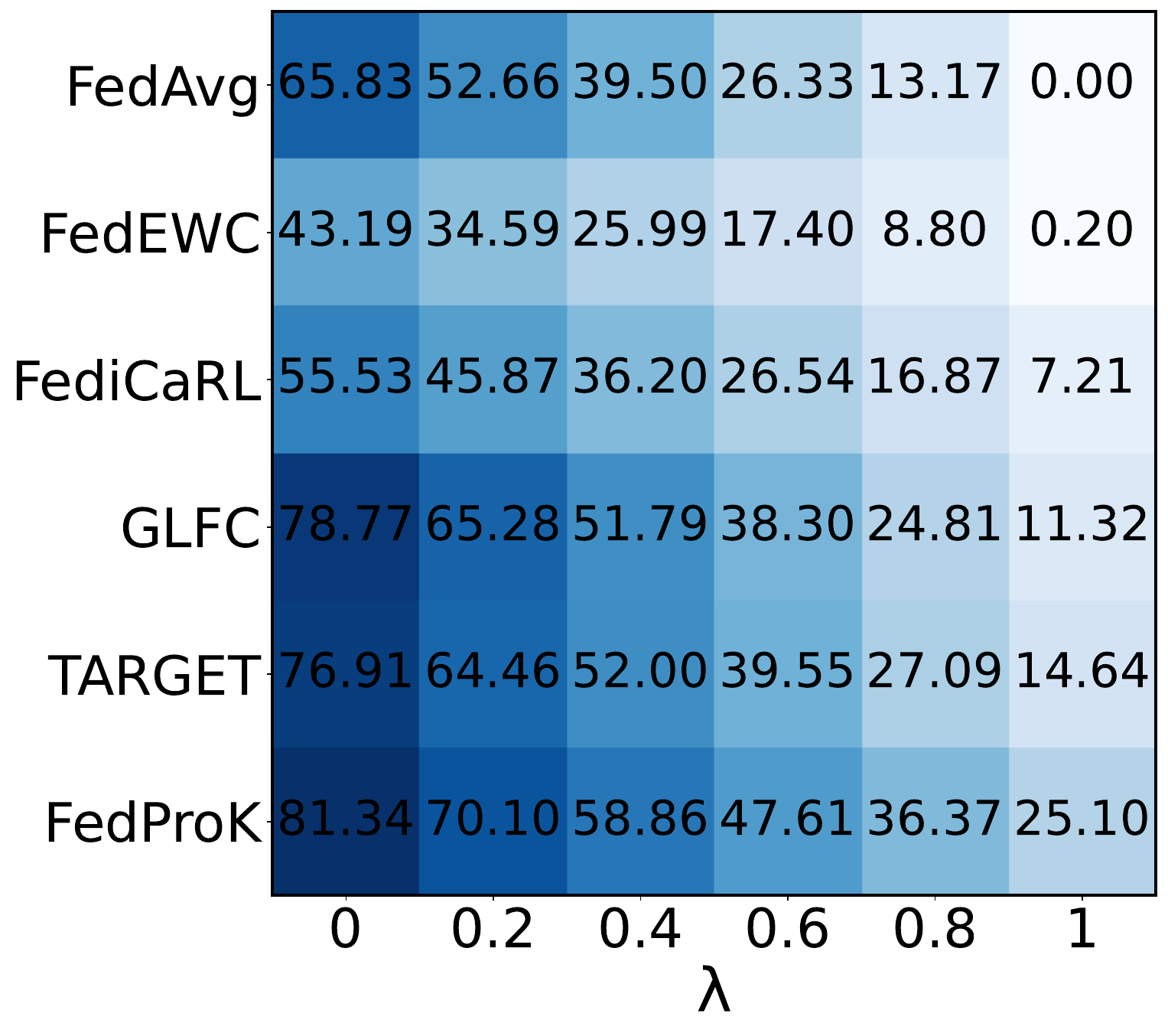}
    \caption{Continual utility.}
    \label{heatmap}
  \end{subfigure}
  \caption{Continual utility of FedProK and baselines on CIFAR-100 with 5 incremental tasks. (a) illustrates the stability and plasticity separately of our model and baselines. (b) illustrates the continual utility as a measurement of the stability-plasticity trade-off of our model and baselines, where $\lambda$ is the hyperparameter. A larger $\lambda$ indicates more emphasis on stability. }
  \label{fig_trustworthy}
\end{figure}

\cref{tradeoff} illustrates the stability and plasticity results separately, indicating that our method achieved state-of-the-art performance on both. Hybrid methods (FedEWC and FediCaRL) completely failed in achieving excellent results on continual utility because of the unacceptably poor stability. FCIL methods (GLFC and TARGET) showed generally better continual utility than hybrid methods. 

Additionally, Continual utility results under different values of $\lambda$ shown in \cref{heatmap} demonstrate that FedProK struck a more favorable balance between stability and plasticity under different $\lambda$, thereby ensuring equilibrium in discriminative capabilities for both previous and new classes.

\subsection{Ablation Study}

As shown in \cref{tab_accuracy}, the effects of each module in our method are validated via ablation studies. Ours-w/oFT and Ours-w/oPKF denote the performance of FedProK without feature translation (FT) on the clients and prototypical knowledge fusion (PKF) on the server. Note that if FT is removed, PKF will not have any effect. Compared with the entire version of FedProK, Ours-w/oFT and Ours-w/oPKF suffer significant performance degradation under all settings. The performance degradation of \textbf{Ours-w/o PKF} supports that the PKF module successfully mitigates data heterogeneity and achieves open-world continual learning by exchanging prototypical knowledge selectively on the server. The performance degradation of \textbf{Ours-w/oFT} supports the overall effectiveness of FedProK in spatial-temporal knowledge transfer by geometrically translating the feature distribution of current classes to that of previous classes and replaying the translated feature sets in subsequent local training. 
\section{Conclusion}
In this work, we formulate a trustworthy FCIL framework with regard to continual utility, efficiency and privacy objectives. We propose FedProK as a solution to the challenges posed by trustworthy FCIL. Specifically, FedProK consists of two components: (1) a client-side feature translation procedure that generates pseudo representations of previous classes to achieve temporal knowledge transfer, and (2) a server-side prototypical knowledge fusion mechanism that shares global prototype lists among participants to achieve spatial knowledge transfer. We conduct experiments in both synchronous and asynchronous settings to demonstrate that our FedProK outperforms the other state-of-the-art methods in terms of accuracy and trustworthiness on common benchmark datasets, which validates the effectiveness of our FedProK in knowledge transfer.

{
    \small
    \bibliographystyle{ieeenat_fullname}
    \bibliography{ref}
}


\end{document}